\documentclass[10pt,journal,compsoc]{IEEEtran}
\usepackage{subfigure}
\usepackage{spconf,amsmath,graphicx,dsfont,color}
\usepackage{amsfonts}
\usepackage{amssymb}
\usepackage{bm}
\usepackage[draft]{hyperref}
\usepackage{algorithm}
\usepackage{algorithmic}
\usepackage{tikz}
\usetikzlibrary{matrix,arrows}

\newcommand{\realset}{\mathbb{R}}
\newcommand{\I}{\mathbf{I}} 
\newcommand{\W}{\mathbf{W}} 
\newcommand{\nsize}{n}  
\newcommand{\graph}{\mathcal{G}} 
\newcommand{\nodes}{\mathsf{V}}
\newcommand{\edges}{\mathsf{E}}
\newcommand{\edge}{\mathsf{e}}
\newcommand{\partition}{\mathsf{P}}
\newcommand{\region}{\mathsf{R}}

\newcommand{\hierarchy}{\mathcal{H}}
\newcommand{\hierarchyset}{\bm{\mathcal{H}}}

\newcommand{\fultra}{\bm{\lambda}}

\newcommand{\tree}{$T$} 
\newcommand{\score}{\texttt{score}}
\newcommand{\train}{T}
\newcommand{\erode}{\varepsilon}
\newcommand{\transformation}{\erode}
\DeclareMathOperator*{\argmin}{arg\,min}
\title{Automatic selection of stochastic watershed hierarchies}
\name{Amin Fehri, Santiago~Velasco-Forero and Fernand Meyer }
\address{
\begin{tabular}{c}
MINES ParisTech, PSL Research University, CMM - Center of Mathematical Morphology \\
\texttt{\{amin.fehri,santiago.velasco,fernand.meyer@mines-paristech.fr\}}
\end{tabular}
}

\begin{document}
%
\maketitle

\begin{abstract}
The segmentation, seen as the association of a partition with an image, is a difficult task. It can be decomposed in two steps: at first, a family of contours associated with a series of nested partitions (or hierarchy) is created and organized, then pertinent contours are extracted. 
A coarser partition is obtained by merging adjacent regions of a finer partition. The strength of a contour is then measured by the level of the hierarchy for which its two adjacent regions merge.
We present an automatic segmentation strategy using a wide range of stochastic watershed hierarchies.
For a given set of homogeneous images, our approach selects automatically the best hierarchy and cut level to perform image simplification given an evaluation score. Experimental results illustrate the advantages of our approach on several real-life images datasets.
\end{abstract}
\section{Introduction}\label{sec:intro}

Image segmentation is the transformation often described as the partitioning of the image domain into a set of meaningful regions according to some pre-specified criteria. It is generally difficult to directly find the pertinent contours in an image, and is thus useful to follow a two-steps strategy: first, we produce a hierarchy, and then extract the meaningful contours out of it. 

To formalize the notion of segmentation, one can see the image as a graph, in which the graph nodes are pixels or regions of the image, and the graph edges link neighbor regions according to a dissimilarity measure. Then cutting all edges of the graph having a valuation superior to a threshold leads to a forest, i.e. a partition of the image. Thus, our goal is to create a partial graph (i.e. a subset of all graph edges) such that its connected components represent the desired segmentation once we return to the image. To do so, we want to find pertinent edges valuations that fully exploit information in the image, so that a cut of the partial graph leads to a segmentation more suitable for further exploitation.
In image processing, such approaches are known as \emph{morphological hierarchical segmentations}. 

Recently, morphological hierarchical segmentations have been studied as a robust way to extract important features of the image \cite{ouzounis12}, to produce robust segmentations of high-dimensional data \cite{gueguen2013local}, to detect important objects in a shape-space \cite{Xu2015}, to find optimal partitions from a given functional\cite{Ravi2013global} as well as regarding computational issues to produce low complexity implementations \cite{najman2013playing}. 

The contributions of this paper are the following: first, we present a method to reevaluate the weights in morphological hierarchies to take more into account the structural information of the images, called stochastic watershed (SWS); second, we show how composing SWS can lead to better results in the sense of the extraction of more significant part of the images; third, we present a workflow to automatically and simultaneously select the best hierarchy of segmentations and the optimal cut-level from a given training set.

The paper is organized as follows. Section \ref{Sec:Back} introduces the notations and the methodology we propose. The experimental setup is described in Section \ref{Sec:Exp}. Section \ref{Sec:Conc} summarizes the conclusions and the future work.

\section{Background and methodology}\label{Sec:Back}

\subsection{Hierarchies and partitions}
\label{ssec:Hierarchies}

To be efficient, we work at two resolutions. The lowest level is the pixel level: the initial image is segmented and a fine partition produced, for instance a set of superpixels \cite{Achanta12,machairas2015waterpixels} or the basins produced by classical watershed algorithm \cite{meyer1990morphological}. We suppose that the fine partition produced by an initial segmentation contains all contours making sense in the image. We define a dissimilarity measure between adjacent tiles of the fine partition. The partition and dissimilarity between adjacent tiles are then modelled as an edge-weighted graph, the \emph{region adjacency graph} (RAG): each node represents a tile of the partition; an edge links two nodes if the corresponding regions are neighbors; the weight of the edge is equal to the dissimilarity between both regions. Working on the graph is much more efficient than working on the image, as there are far less nodes in the graph that there are pixels in the image.

Formally, we define a non oriented graph $\graph=(\nodes,\edges,\W)$ as containing a set $\nodes$ of nodes or vertices, a set $\edges$ of edges, an edge being a pair of nodes, and an edges weight function $\W: \edges \to \realset^{+}$. The edge linking the nodes $p$ and $q$ is designated by
$e_{pq}$. The partial graph associated with the edges $\edges_{0} \subset \edges$ is $\graph_{0} = (\nodes,\edges_{0},\W)$.

A \emph{path} $\pi$ is a sequence of nodes and edges: for example $\pi = \{p,e_{pt},t,e_{ts},s\}$ is a path linking the nodes $p$ and $s$.
A \emph{connected subgraph} is a subgraph where each pair of nodes is connected by a path. A \emph{cycle} is a path whose extremities coincide. A \emph{tree} is a connected graph without cycle.
A \emph{spanning tree} is a tree containing all nodes. A \emph{minimum spanning tree} (MST) is a spanning tree with minimal possible weight, obtained for example using the Boruvka algorithm. A \emph{forest} is a collection of trees.

\begin{figure}
\includegraphics[width=.32 \columnwidth]{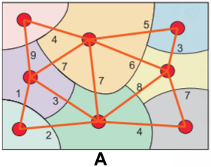}
\includegraphics[width=.32 \columnwidth]{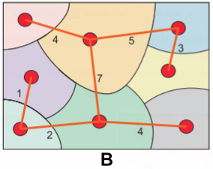}
\includegraphics[width=.32 \columnwidth]{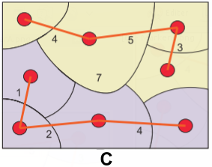}
\caption{\textbf{A}: a partition represented by an edge weighed graph; \textbf{B}: a minimum spanning tree of the graph; \textbf{C}: two connected subtrees, obtained by cutting all edges with a weight above 6, and the underlying segmentation indicated by the two colors.}\label{Fig:MST}
\end{figure}

Cutting all edges having valuations superior to a threshold $\lambda$ in a RAG leads to a forest in which each tree represents a region in the image. 
By giving the same labels to all nodes of the same trees, one obtains a \emph{segmentation} of the image. So that the more edges are cut, the more the subtrees are subdivised and the finer the segmentation. An illustration of the passage from a RAG to a partition/segmentation can be found in Figure \ref{Fig:MST}. In this regard, the obtained result is the same whether we work with the graph or with an MST of the graph. For the sake of simplicity, we thus work only with the MST in the sequel.
In the MST, cutting the $n$ highest edges leads to a \emph{minimal spanning forest} (MSF) of $n+1$ trees. Cutting another edge subdivides one of these $n+1$ trees in two subtrees.

\begin{figure}
\subfigure[$\I$]{\includegraphics[width=.485 \columnwidth]{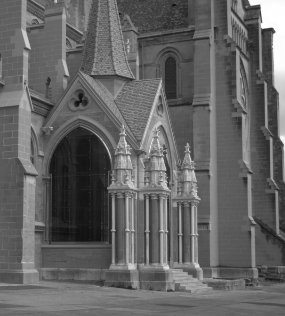}\label{imageForms}}
\subfigure[$\fultra_{\texttt{Grad}}$]{\includegraphics[width=.485 \columnwidth]{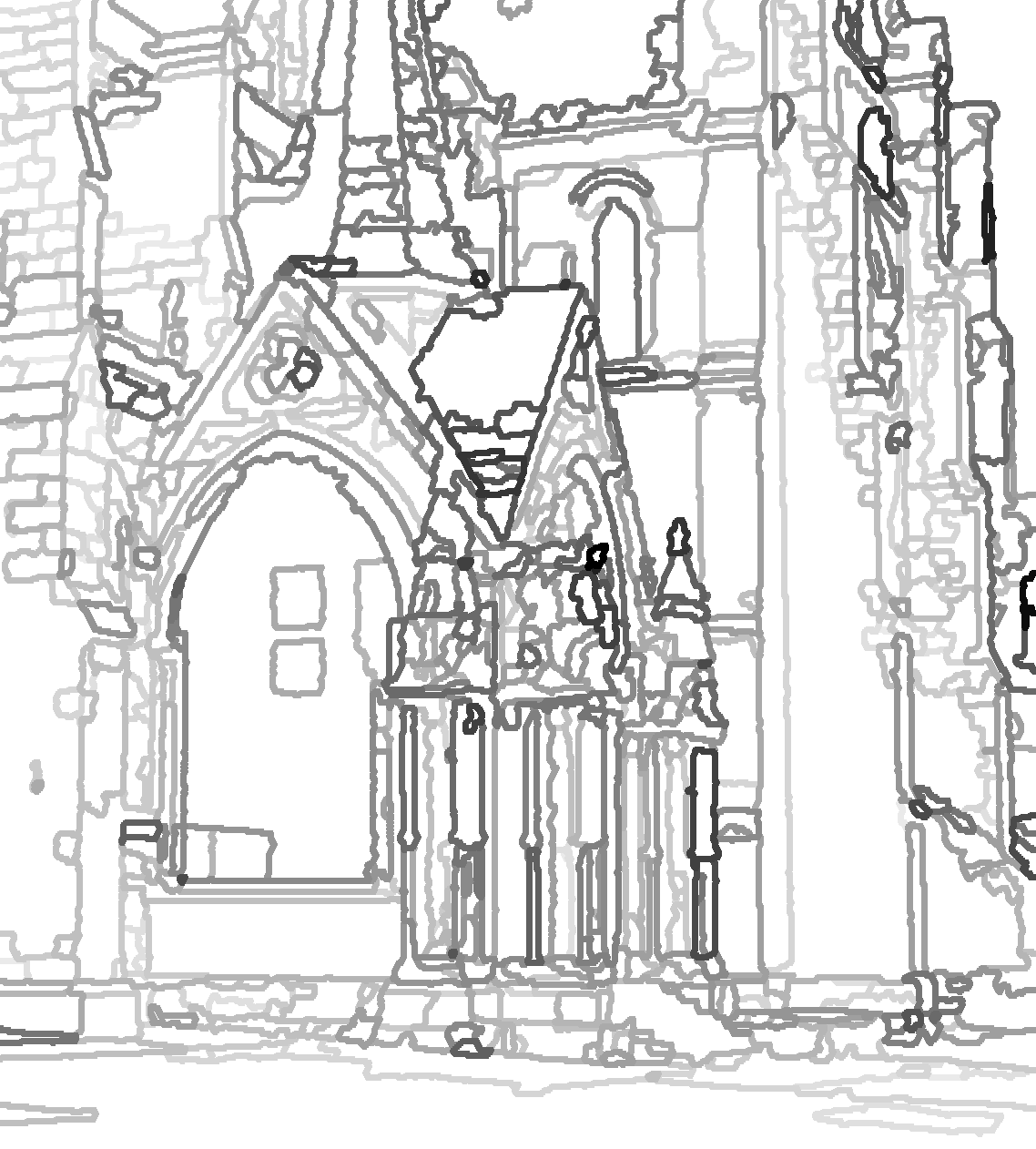}\label{imSGrad}}\\
\subfigure[$\fultra_{\texttt{SSurf}}(\fultra_{\texttt{Grad}})$]{\includegraphics[width=.485 \columnwidth]{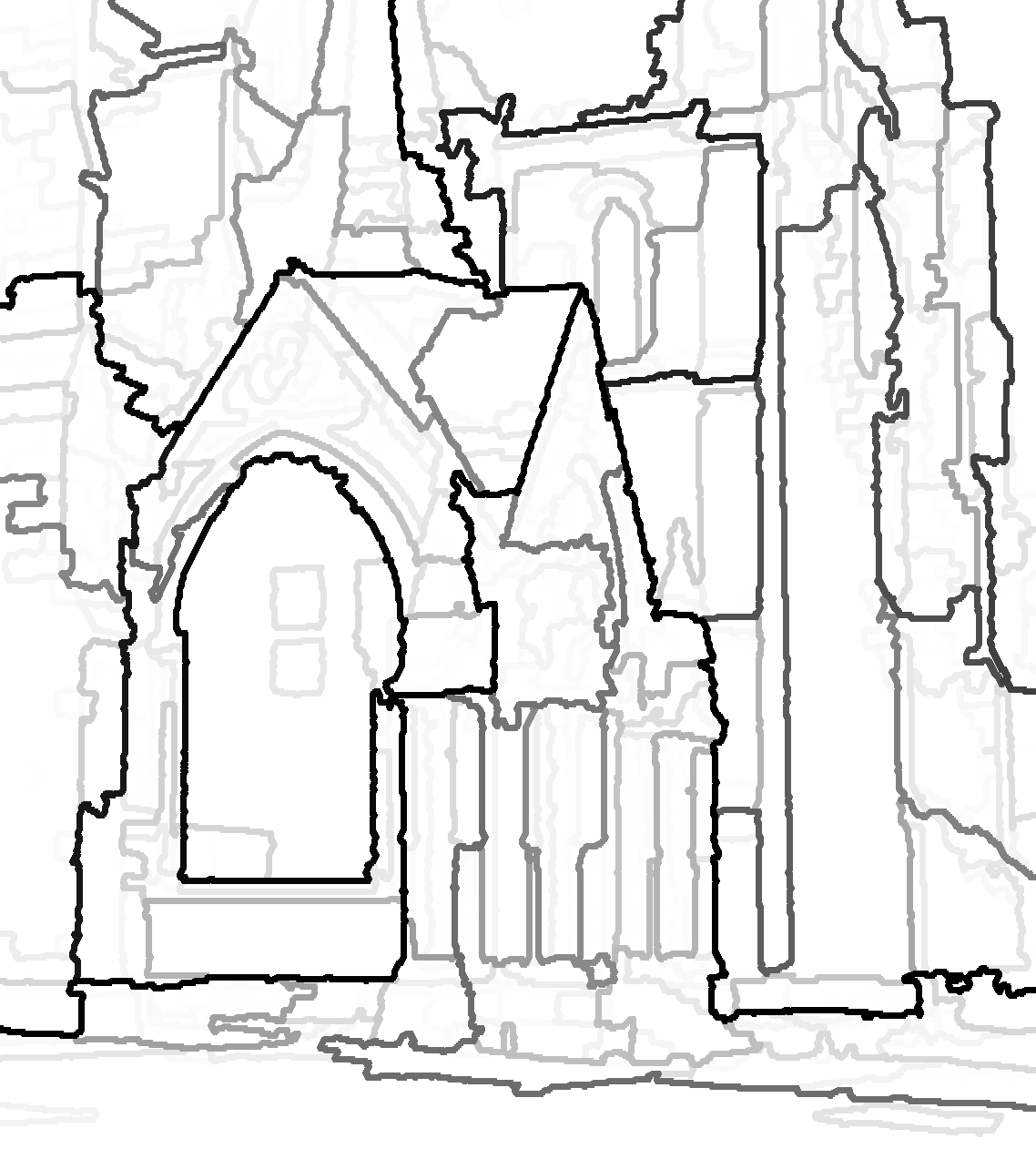}\label{imSSurf}}
\subfigure[$\fultra_{\texttt{SVol}}(\fultra_{\texttt{Grad}})$]{\includegraphics[width=.485 \columnwidth]{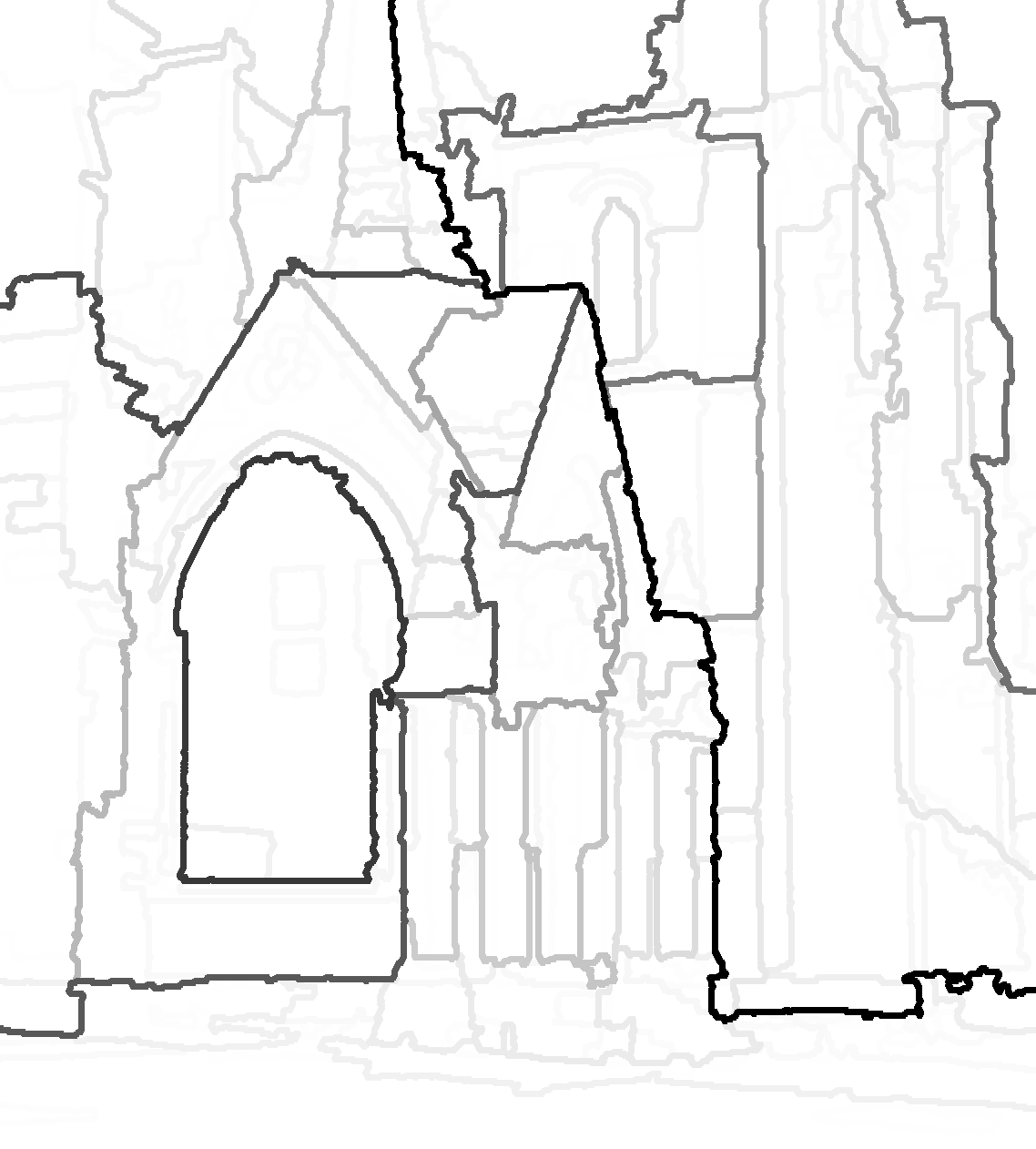}\label{imSVol}}\\
\subfigure[$\fultra_{(\texttt{SSurf},\transformation_{\rightarrow})}(\fultra_{\texttt{Grad}})$]{\includegraphics[width=.485 \columnwidth]{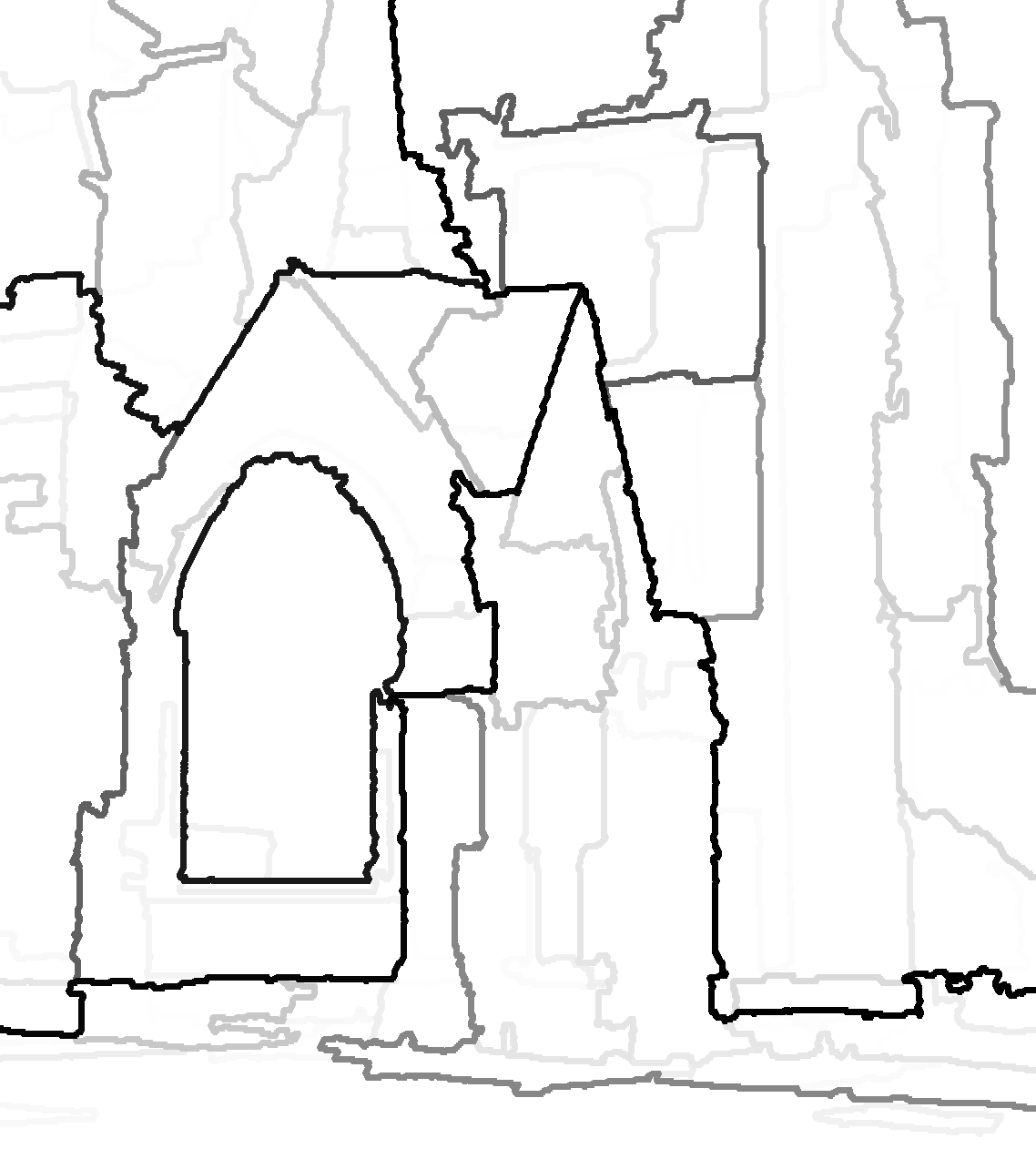}\label{imSSurfHoriz}}
\subfigure[$\fultra_{(\texttt{SSurf},\transformation_{\uparrow})}(\fultra_{\texttt{Grad}})$]{\includegraphics[width=.485 \columnwidth]{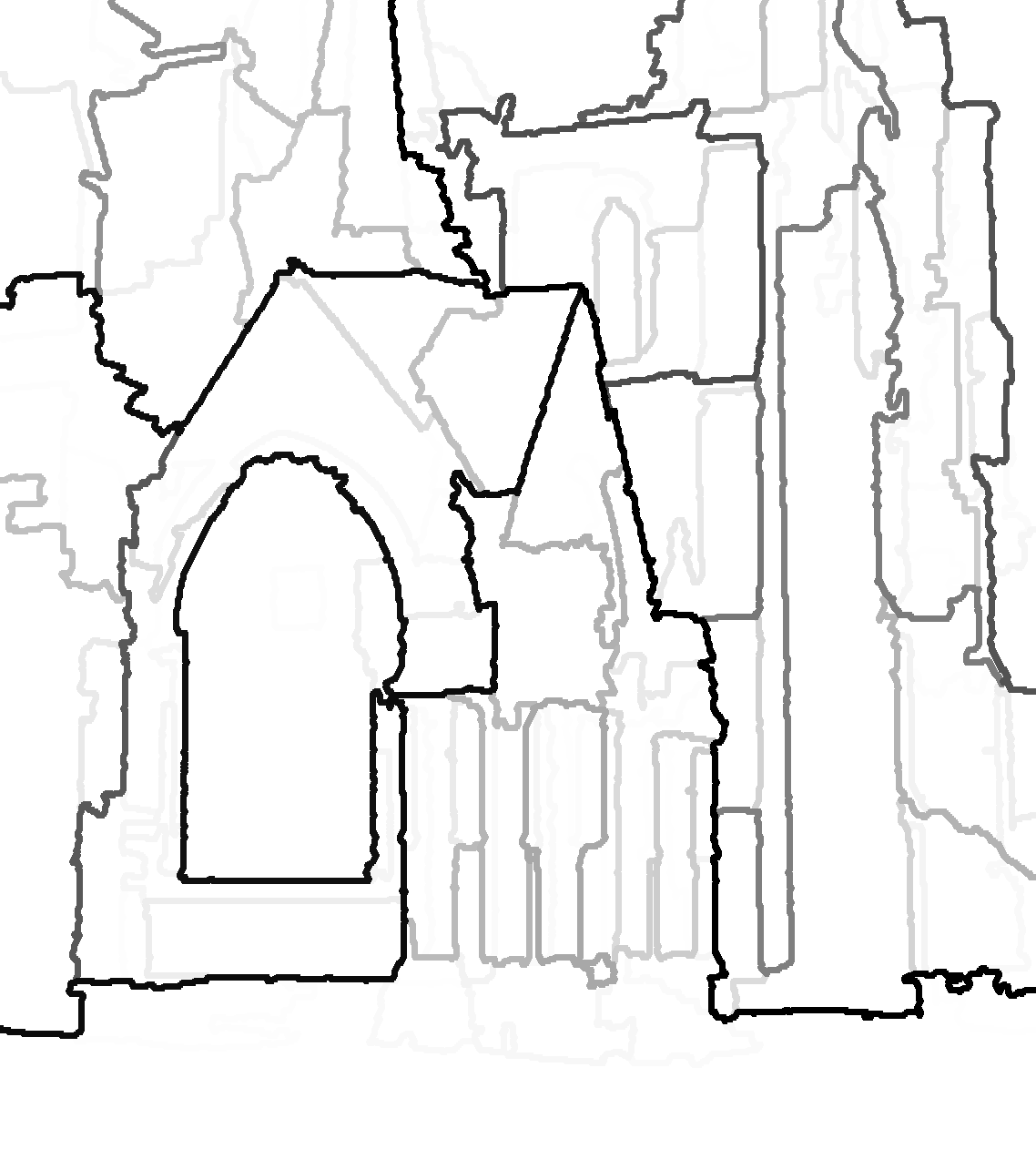}\label{imSSurfVert}}\\
\caption{The indexed hierarchies are illustrated by their saliency function $\fultra$. $\transformation_{\circ}$ denotes the erosion with a circle of size four as structuring element. $\transformation_{\rightarrow}$ denotes the erosion with a horizontal segment of size four as structuring element. $\transformation_{\uparrow}$ denotes the erosion with a vertical segment of size fifteen as structuring element}
\vspace{-1em}
\label{FigSaliences}
\end{figure}

Cutting edges by decreasing valuations thus gives an \emph{indexed hierarchy of partitions} $(\hierarchy,\fultra)$, with $\hierarchy$ a \emph{hierarchy of partitions} i.e. a chain of nested partitions $\hierarchy=\{\partition_0, \partition_1,\ldots, \partition_\nsize| \forall j,k, \quad 0 \quad \leq j\leq k\leq \nsize \Rightarrow \partition_j \sqsubseteq \partition_k\}$, with $\partition_\nsize$ the single-region partition and $\partition_0$ the finest partition on the image, and $\fultra: \hierarchy \to \realset^+$  being an increasing map taking its values into the decreasing cut valuations, such that for two nested partitions $\partition \subset \partition'$, we have $\fultra(\partition) < \fultra(\partition')$. This increasing map allows us to value each contour according to the cut level of the hierarchy for which it disappears: this is the \emph{saliency} of the contour, and the higher the saliency, the strongest the contour, as illustrated in Figure \ref{FigSaliences}.

The quality of the hierarchy, i.e. the pertinence of the obtained regions at various levels of it, depends on the dissimilarity used. If the dissimilarity reflects only a local contrast, the most salient regions in the image are the small contrasted ones, as illustrated in Figure \ref{FigSaliences}b. In the following section, we explain how to construct more pertinent and informative dissimilarities. 

\subsection{Marker-based segmentation}
\label{ssec:MBS}

\begin{figure}
\includegraphics[width=.99 \columnwidth]{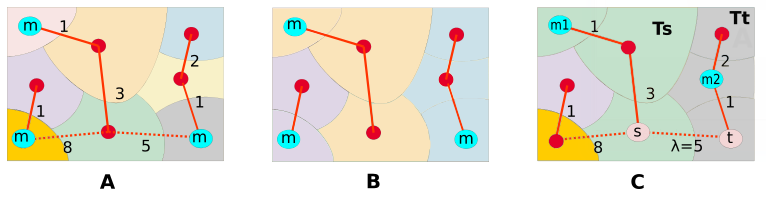}
\caption{\textbf{A}: a MST with markers - highest edges on the path linking them are cut; \textbf{B}: the corresponding partition; \textbf{C}: considering two markers $m1$ and $m2$ falling into the regions spanned by the two subtrees $\tree_{s}$ and $\tree_{t}$ obtained when cutting an edge $\edge_{st}$ on the MST, the highest edge on the path linking them is indeed $\edge_{st}$.}\label{Fig:MarkersSeg}
\end{figure}

To do so, one can select a node in each region or object of interest that will serve as a root in each wanted tree. We then construct an MSF in which each tree takes root in one of the selected nodes. The roots are also called \emph{markers} and this process referred to as \emph{marker-based segmentation}. The final result is obtained by suppressing, for each pair of markers, the highest edge on the unique path on the MST linking them. This is illustrated in Figures \ref{Fig:MarkersSeg}A,B. 

Let us consider an edge $\edge_{st}$ of the MST of weight $\lambda$, and let us examine what markers lead to a selection of this edge.
If we put at least one marker within the domains spanned by each of these trees, then the highest edge on the unique path on the MST linking them is indeed $\edge_{st}$, which is thus cut in the associated segmentation. This mechanism can be observed on Figure \ref{Fig:MarkersSeg}C), in which we can see that the highest edge on the path linking two markers $m_{1}$ and $m_{2}$ put respectively in $\tree_{s}$ and $\tree_{t}$ is indeed $\edge_{st}$. We say that a root is chosen in $\tree_{s}$ if at least one marker lies within its corresponding region in the image obtained by replacing each node of the tree by the region it represents.

Note that this process is robust to the choice of markers, since the selection of any node of a region as a root leads to a segmentation of this region.

\subsection{Stochastic Watershed Hierarchies}
\label{ssec:SWS}

Rather than using deterministic markers, one can use random markers following a given distribution and thus generate random MSF. Then, one can assign to each edge of the MST the probability for it to be cut accordingly, which corresponds to the probability of appearance of the underlying contour. 
Depending on the distribution of markers used, the obtained segmentation can be very variable.  

This idea finds its source in the stochastic watershed presented by Angulo in \cite{angulo2007stochastic}. If we see the image as a topographic relief, flooding this image leads to watershed lines, i.e. to a segmentation. By spreading random flooding sources multiple times and flooding the image accordingly, one can characterize each contour of the image by its frequency of appearance in the associated segmentations.
We obtain similar results with computations made directly on graphs following the nomenclature presented in section \ref{ssec:Hierarchies}. 

Moreover, we can define many ways to generate markers, depending on the probability law used to implant them, but also on their sizes or shapes if they are non-points. Each particular mechanism favors the emergence of a certain type of regions to the detriment of others. Keen readers are invited to refer to \cite{Meyer15} for more details. We take advantage of this versatility to compute new dissimilarity values that are more linked with the semantic information present in the image.

To summarize, departing from a tree which valuations are those of the initial graph, one obtains a tree with identical structure but different edges valuations. To go further, this new tree can then be used as a departure point for a similar construction but based upon another type of markers and another random distribution law of them. We call this process \emph{composition} of hierarchies. 

Note that to each pair of markers and distribution corresponds a new hierarchy of partitions. So that we now wonder, for a given segmentation task, which one of these hierarchies approaches us more to our final purpose, i.e. to obtain a representative partition of the analyzed scene.

The subject of our paper is to present, for given segmentation task and images set, a procedure enabling us to automatically select pertinent hierarchy and cut level in order to get an adequate segmentation.

In our paper, we get a partition from a hierarchy by thresholding the highest edges as described above. A marker-based segmentation can be applied to the result as well and other approaches have been proposed, such as interactive segmentation or energy minimization techniques \cite{Ravi2013global}. 

\subsection{Finding a well-suited hierarchy and cut level from a training set}

We have now many ways to interrogate our images using different combination of hierarchies. A segmentation of the image is given by choosing a level of a hierarchy applied to this image. Although, it is hard to know, for a given set of images, which hierarchy and which level of this hierarchy would give good results regarding the segmentation task. It is common that images to segment share similar properties, due to their nature or to the tools allowing us to visualize them, as for example for cells images in microscopy, or bones and tissues images in radiography. To facilitate the obtention of a satisfying segmentation, it is in our interest to find a hierarchy that takes into account these shared properties amongst each images collection. In a tailor approach, we thus propose a methodology to automatically select a pertinent hierarchy and a good cut level of it for a given set of homogeneous images, so that a suitable segmentation can be obtained for a new image of the same kind without effort.  

Let us say we have at our disposal a $\score( \I , (\hierarchy,\lambda))$ to judge the quality of a segmentation $(\hierarchy,\lambda)$ obtained for an image $\I$. Note that $(\hierarchy,\lambda)$ is the partition obtained after setting the value of the indexed hierarchy $(\hierarchy,\fultra)$ to $\lambda$. Thus, we would like to find the best hierarchy and the best cut level $\lambda$ according to the score evaluated on a training set of images.    
Formally, given a training set $\train=\{\I_1,\ldots,\I_{|\train|}\}$ and a set of indexed hierarchies $\hierarchyset=\{(\hierarchy_1,\fultra_1),(\hierarchy_2,\fultra_2),\ldots,(\hierarchy_{|\hierarchyset|},\fultra_{|\hierarchyset|}) \}$, we are interested in finding the hierarchy $\hierarchy$ and cut level $\lambda$ that minimize the score for the training set, i.e., 
\begin{equation}
(\hierarchy^*,\lambda^*):=\argmin_{(\hierarchy,\lambda \in \fultra) \in \hierarchyset}  \sum_{i=1}^{T} \score(\I_i,(\hierarchy,\lambda)).
\label{modelH}
\end{equation}

Let us consider a set of homegenous images, that we subdivide into training and testing subsets, and a set of indexed hierarchies $\hierarchyset$ (possibly composition of hierarchies as in Section \ref{ssec:Hierarchies}). We take advantage of the low computational cost of our approach (only involving updates in the MST) to find the optimal hierarchy in \eqref{modelH} by an exhaustive search on the training subset. We call this learned hierarchy the \emph{model} hierarchy. 

To test its pertinence, we apply it on the testing subset. For each test image $\I$, we can also find by an exhaustive search as well the best possible hierarchy and cut level,  designated as the \emph{oracle}:
\begin{equation}
(\hierarchy^{\texttt{oracle}},\lambda^{\texttt{oracle}}):=\argmin_{(\hierarchy,\lambda \in \fultra) \in \hierarchyset} \score(\I,(\hierarchy,\lambda)).
\label{oracleH}
\end{equation}

One can say we have effectively found a good model hierarchy for the set of images if the difference between the scores obtained for the model \ref{modelH} and the oracle \ref{oracleH} is on average low on the test subset.

\section{Experimental results}\label{Sec:Exp}
We consider in this work all combinations up to depth two of the following SWS hierarchies: watershed hierarchy (or gradient based), surface-based, volume-based, surface-based after erosion and volume-based after erosion.

\subsection{Type of Scores}
The model proposed in the previous section is suitable for any score that we want to minimize (or maximize) in order to get a good segmentation. To test this model, we use two different scores.

The first score used is a Mumford-Shah score \cite{pock2009algorithm}, so that the problem we want to solve is an energy minimization problem. This score contains two terms, a data fidelity term and a regularization term: by climbing in the hierarchy towards coarser levels, the value of the first term increases and the value of the second one decreases. Both terms are linked by a scale parameter.
It has this form:
\begin{equation}
\texttt{MS}(\pi = (\I,\hierarchy,\lambda)) = \sum_{\region_{i} \in \pi} \texttt{var} (\region_{i}) + s C(\pi) ,   
\end{equation}
where $\texttt{var}(\region_{i})$ represents the total variance of the image in the region $\region_{i}$ of the partition $\pi = (\I,\hierarchy,\lambda)$, $C_{\pi}$ represents the length of the contours present in the partition $\pi$, and $s$ is a scale parameter that allows to have a trade-off between data fidelity and a simplification of the image.

The second score that we used is a new metric called ``weighted human disagreement rate"(WHDR), introduced in \cite{bell14intrinsic} to evaluate intrinsic decomposition results. It is associated with the large-scale public database, Intrinsic Images in the Wild (IIW), built in \cite{bell14intrinsic}, and composed of 5230 manually annotated images of complex real indoor scenes. 
WHDR measures the level of agreement between the judgements made by algorithms being evaluated and those of humans. The cut of the hierarchy allows us to obtain a kind of reflectance image for each test image and so to use this score to evaluate it. The WHDR varies between 0 and 1, being close to 0 when the reflectance image is consistent with human judgment, and close to 1 otherwise.

\subsection{Results}

\begin{figure}
\subfigure[Some examples in the training set]{\includegraphics[width=.245 \columnwidth]{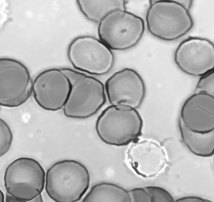}\;\includegraphics[width=.245 \columnwidth]{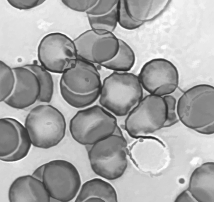}\;\includegraphics[width=.245 \columnwidth]{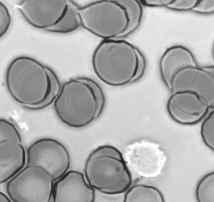}\;\includegraphics[width=.245 \columnwidth]{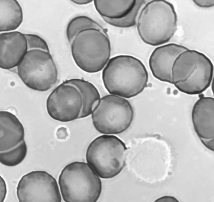}}\\
\subfigure[$\I$]{\includegraphics[width=.325 \columnwidth]{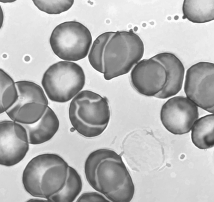}}
\subfigure[\texttt{model}]{\includegraphics[width=.325 \columnwidth]{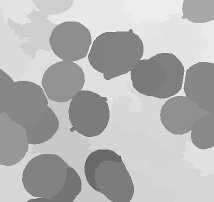}}
\subfigure[\texttt{oracle}]{\includegraphics[width=.325 \columnwidth]{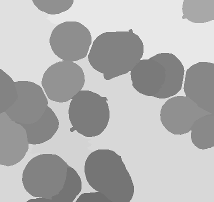}}\\
\subfigure[$\I$]{\includegraphics[width=.325 \columnwidth]{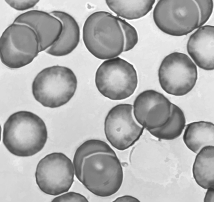}}
\subfigure[\texttt{model}]{\includegraphics[width=.325\columnwidth]{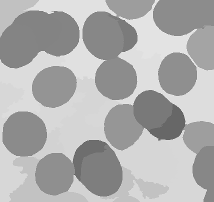}}
\subfigure[\texttt{oracle}]{\includegraphics[width=.325 \columnwidth]{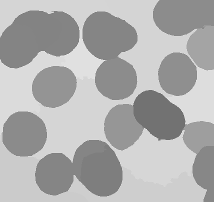}}\\
\subfigure[$\I$]{\includegraphics[width=.325 \columnwidth]{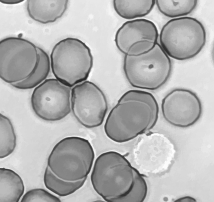}}
\subfigure[\texttt{model}]{\includegraphics[width=.325 \columnwidth]{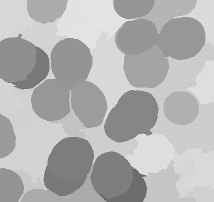}}
\subfigure[\texttt{oracle}]{\includegraphics[width=.325 \columnwidth]{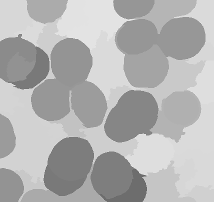}} 
\caption{Results on some examples of cells, for a Mumford-Shah score with a scale parameter $s$ = 1.168. (b),(e),(h) are images from the testing set, (c),(f),(i) the model segmentations and (d),(g),(j) the oracle segmentations.}\label{MSExamples}
\vspace{-1.5em}
\end{figure}

\begin{figure}
\subfigure[Some examples in the training set]{\includegraphics[width=.245 \columnwidth]{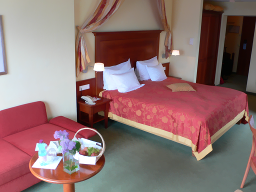}\;\includegraphics[width=.245 \columnwidth]{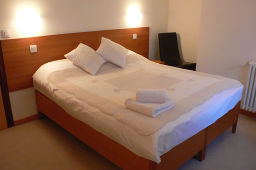}\;\includegraphics[width=.245 \columnwidth]{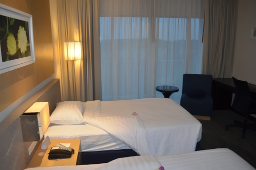}\;\includegraphics[width=.325 \columnwidth]{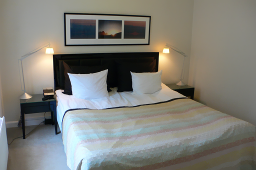}}\\
\subfigure[$\I$]{\includegraphics[width=.325 \columnwidth]{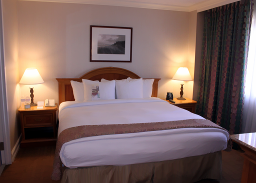}}
\subfigure[\texttt{model}]{\includegraphics[width=.325 \columnwidth]{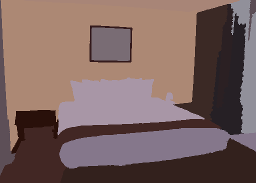}}
\subfigure[\texttt{oracle}]{\includegraphics[width=.325 \columnwidth]{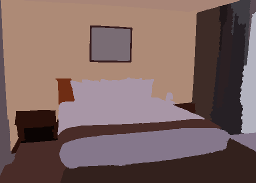}}\\
\subfigure[$\I$]{\includegraphics[width=.325 \columnwidth]{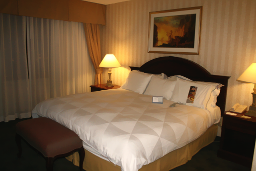}}
\subfigure[\texttt{model}]{\includegraphics[width=.325\columnwidth]{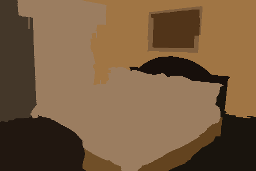}}
\subfigure[\texttt{oracle}]{\includegraphics[width=.325 \columnwidth]{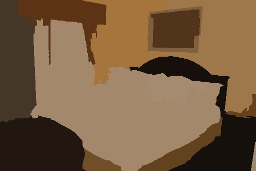}}\\
\subfigure[$\I$]{\includegraphics[width=.325 \columnwidth]{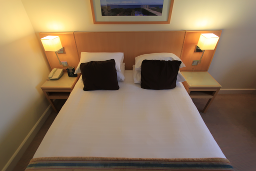}}
\subfigure[\texttt{model}]{\includegraphics[width=.325 \columnwidth]{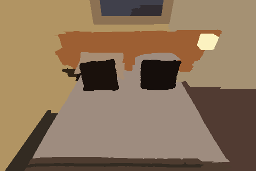}}
\subfigure[\texttt{oracle}]{\includegraphics[width=.325 \columnwidth]{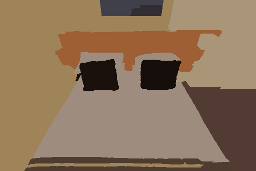}} 
\caption{Results on some examples of Intrinsic Images in the Wild, for a WHDR score.(b),(e),(h) are images from the testing set, (c),(f),(i) the model segmentations and (d),(g),(j) the oracle segmentations.}\label{WHDRExamples}
\vspace{-1.5em}
\end{figure}

In a first approach, we tested our strategy with a set of cells images for the Mumford-Shah score, and with homogeneous subsets of images from the IIW database, of bedrooms, bathrooms and people. Some visual results can be found on Figs.\ref{MSExamples} and \ref{WHDRExamples}.

For each set of images, we train our system on a subset to learn the best hierarchy among any hierarchy or combination of two hierarchies presented before, which provides us with a model hierarchy. Then, for each image of the test subset, we compute the optimal hierarchy for this precise image, that is the oracle hierarchy, the score attached to it, as long as the score given by the model hierarchy on this test image. 
A summary of the results for the WHDR score is given in Table \ref{ResultsTab}.

%


\begin{table}[]
\centering
\label{my-label}
\begin{tabular}{|l|l|l|l|l|}
\textbf{Database} & \textbf{$\mu$(WHDR$^{\texttt{oracle}}$)} & \textbf{$\mu$(WHDR$^{\texttt{model}}$)} &	\textbf{$\mu$(\texttt{error})} & \textbf{$\sigma$(\texttt{error})} \\
Bathrooms         & 0.154  & 0.178 & 0.024  &  0.025   \\
People            &  0.282  &   0.133  &  0.148 &	0.093 \\
Bedrooms          & 0.125  &    0.237 &  0.112  &  0.107                 
\end{tabular}
\caption{Mean and standard deviation of the error between $\texttt{oracle}$ and $\texttt{model}$, i.e. the difference between WHDR$^{\texttt{model}}$ and WHDR$^{\texttt{oracle}}$, and averages of the scores for the oracle and model for the different test databases and the WHDR measure.}
\label{ResultsTab}
\end{table}

Furthermore, we can have insights about why a hierarchy has been chosen for a given set of images.   
For the cells images, the model hierarchy found by the algorithm is a composition of a volumic SWS followed by a surfacic SWS. We can interpret it as a first step eliminating non pertinent objects with a trade-off between area and gradient, and a second pass emphasizing the cells based on their surface, since it is often of the same order. For the IIW images, the hierarchy selected is a composition of volumic SWS, with different sizes and orientations for structuring elements depending of the dataset. One can interpret that the type of objects usually present in the image differs regarding the scenes, and thus the adaptive hierarchies depend on the forms found in the images. Surfacic SWS are not very pertinent here since there is a wide variety of objects of different sizes in the images.         

\section{Conclusions}\label{Sec:Conc}
In this paper we have presented a novel approach to compose hierarchies of segmentations. The proposed workflow has been evaluated for a difficult task: the obtention of the best hierarchy and cut to perform image simplification given an evaluation score. To go further, several enhancements of the system are conceivable. One could use combinations of hierarchies longer than two. The first scores here used may also be replaced by other ones, more adapted for a given task, for example a score to use for interactive segmentation translating the difficulty for the user to get the desired result from the obtained segmentation. One could also imagine using a similar methodology to characterize sets of homogeneous images.

\bibliographystyle{IEEEbib}
\begin{small}
\bibliography{BibSegmentation}
\end{small}
\end{document}